\documentclass[mlmain]{jmlr}
\usepackage{longtable}% for long tables

\usepackage{booktabs}
\usepackage[load-configurations=version-1]{siunitx} % newer version
\theorembodyfont{\upshape}
\theoremheaderfont{\scshape}
\theorempostheader{:}
\theoremsep{\newline}

\jmlrvolume{}
\firstpageno{1}
\editors{}

\jmlryear{2026}
\jmlrworkshop{Symmetry and Geometry in Neural Representations}

\editor{Editor's name}
\title[Receptive Fields and Neuronal Expressivity]{The Computational Value of Sensory-Aligned Receptive\titlebreak Fields Depends on Neuronal Expressivity}

  \author{\Name{Agnese Adorante} \Email{agnese.adorante@student.uni-tuebingen.de}\\
  \Name{Aaron Spieler} \Email{aaron.spieler@tuebingen.mpg.de}\\
  \Name{Anna Levina} \Email{anna.levina@uni-tuebingen.de}\\
   \addr University of T\"ubingen, Germany}

\begin{document}
\maketitle

\begin{abstract}
Biological sensory neurons have selective receptive fields organized along meaningful stimulus coordinates, such as frequency, motion direction, or retinotopic position. Such structure may arise from efficient coding and biological constraints on activity, connectivity, and wiring, as computational studies of simple neurons have shown across modalities. This raises a question: do structured receptive fields confer a computational advantage beyond resource efficiency itself, and does this advantage persist when individual neurons are highly expressive?
We address this question in recurrent networks of Expressive Leaky Memory neurons, where we can independently vary neuronal complexity and the organization of feed-forward receptive fields. Across auditory and event-based visual classification tasks, receptive fields aligned with a task-relevant sensory coordinate improve test accuracy relative to budget-matched random receptive fields. This advantage disappears when sensory coordinates are scrambled, or when receptive fields follow task-irrelevant coordinates, showing that the benefit comes from alignment with task geometry rather than restricted connectivity alone. Increasing neuronal complexity reduces the performance advantage of structured receptive fields. Finally, generic synaptic sparsity regularization induces input selectivity and partially recovers performance, but remains below explicitly structured receptive fields, suggesting that sparsity alone may not fully capture the computational benefit of task-aligned receptive fields. Together, our results show that appropriate receptive fields can serve as a computational prior beyond sparsity itself, and that their value depends on the computational expressivity of individual neurons.
\end{abstract}

%%%%%%%%%%%%%%%%%%%%%%%%%%%%%%%%%%%%%%%%%%%%%%%%%
\section{Introduction}
\label{sec:intro}
%%%%%%%%%%%%%%%%%%%%%%%%%%%%%%%%%%%%%%%%%%%%%%%%%
Sensory neurons are selective for restricted regions or features of their stimulus spaces. In audition, neurons respond preferentially to particular frequency ranges \citep{merzenich1975}; in vision, retinal and cortical neurons respond to restricted regions of visual space \citep{hubel1962}; and neurons in the visual cortex can additionally be tuned to features such as motion direction \citep{hubel1968,albright1984}. Such selectivity is commonly described through \textit{receptive fields}, which characterize the regions or features of sensory space over which stimuli modulate a neuron's response. 

Computational studies across sensory modalities have identified routes for the emergence of receptive-field structure, often using optimization for efficient representation and resource use. In vision, sparse coding of natural images results in localized filters resembling primary visual cortex receptive fields \citep{olshausen1996}, while related efficient-coding objectives recover other characteristic forms of visual selectivity \citep{bell1997,karklin2011}. In audition, efficient representations of natural sounds similarly produce frequency-selective and spectrotemporal filters resembling those in the auditory pathway \citep{smith2006}. More explicitly, biological constraints such as metabolic costs, limited activity, and the spatial cost of neuronal wiring can push the system toward selective organization \citep{attwell2001,chklovskii2002}. Thus, receptive-field structure can emerge as an efficient solution for representing sensory inputs under constraints. This, however, leaves a separate functional question: is such structure only a consequence of resource optimization, or does it confer a computational advantage beyond the constraints that give rise to it?

Optimal input sampling, additionally, may depend on the neuron's computational capacity. Biological neurons are themselves rich dynamical systems, capable of nonlinear integration and computation over multiple timescales \citep{beniaguev2021,  spieler2024elm}. More generally, computational load can be distributed between individual neurons and the complexity of the network in which they are embedded \citep{spieler2026scaling}. It remains unclear how the computational value of receptive-field structure should depend on the complexity of the neurons themselves.

Whether receptive-field restriction provides a useful computational prior is also a natural question in machine learning. Architectural inductive biases exploit known structure in the input to restrict what must be learned from data \citep{bronstein2021}. Convolutional networks exploit retinotopic geometry through local receptive fields, but combine locality with weight sharing and translation equivariance \citep{lecun1998}. Their success demonstrated the value of a broader geometric prior; however, as artificial units become more expressive, it is unclear how much benefit remains in explicitly imposing sensory geometry rather than allowing the model to learn the relevant selectivity. From both the neuroscience and machine-learning perspectives, this leads to the same question: how should computation be divided between the structure of a neuron's inputs and the computational capacity of the neuron itself?

We address this question using recurrent networks of \textit{Expressive Leaky Memory} (ELM) neurons \citep{spieler2024elm,spieler2026scaling}, whose intrinsic computational capacity can be varied independently of their feed-forward connectivity. We operationalize receptive-field structure by restricting each neuron's feed-forward inputs along a sensory coordinate, and compare receptive-field networks with controls receiving the same number of inputs without any sensory geometry. We use modality-specific receptive-field organization, employing frequency structure for an auditory task, motion-direction structure for gesture recognition, and retinotopic structure for event-based object recognition. Task-aligned receptive fields consistently improve generalization, and the advantage disappears when the sensory coordinate is scrambled or when structure is imposed along a task-irrelevant coordinate. Increasing neuronal expressivity reduces the advantage. Finally, generic synaptic sparsity regularization induces input selectivity and partially recovers performance, but remains below explicitly structured receptive fields, suggesting that sparsity pressure alone may not fully recover their computational benefit. Together, these results show that geometry-aligned receptive fields provide a useful computational prior on the neurons that receive them. However, the value of this prior depends on the expressivity of those neurons.
%%%%%%%%%%%%%%%%%%%%%%%%%%%%%%%%%%%%%%%%%%%%%%%%%
\section{Related Work}
\label{sec:related}
%%%%%%%%%%%%%%%%%%%%%%%%%%%%%%%%%%%%%%%%%%%%%%%%%
Prior work in neuroscience and machine learning has studied receptive-field structure either as a consequence of resource constraints or as an inductive bias. In both literatures, the value of such structure has been shown to depend on the computational capacity available. 

\paragraph{In neuroscience: Resource constraints shape receptive fields.}
Receptive fields have been explained as solutions to optimization problems under biological constraints. \citet{ocko2018retina} optimize a retinal model to maximize information transmission under a firing-rate budget and recover receptive fields resembling those of major primate retinal ganglion cell types. \citet{lindsey2019retina} show that the resulting structure also depends on downstream network depth: a shallow network produces general-purpose filters, whereas a deeper network produces task-specialized ones. A similar dependence has been reported for color processing \citep{harris2021opponency}.

These studies vary bandwidth or network depth. We instead ask how the value of structured input connectivity changes with the expressivity of individual units, motivated by evidence that single neurons can implement substantial computation \citep{beniaguev2021,spieler2026scaling}.

\paragraph{In machine learning: locality as an inductive bias.}
Convolutional networks combine local connectivity with weight sharing \citep{lecun1998,bronstein2021}, making their individual contributions difficult to disentangle. \citet{elsayed2020lowrank} interpolate between these properties. \citet{lahoti2024locality} compare fully connected, locally connected, and convolutional networks and find that stronger structural priors reduce the amount of data needed to reach a given performance, but locally connected networks still underperform convolutional ones at scale.

Further work shows that weaker locality priors can perform comparably when models have sufficient capacity and data: vision transformers can match convolutional networks at scale \citep{dosovitskiy2021vit}, while soft locality priors improve sample efficiency but can be relaxed during training \citep{dascoli2021convit}. These studies therefore suggest a capacity-dependent value of architectural priors, but vary capacity at the network or training level rather than within individual units.
\paragraph{Selectivity can emerge without prescribed structure.}
Selectivity can arise without being explicitly specified, under generic constraints such as sparsity \citep{olshausen1996,smith2006,neyshabur2020convolutions,ingrosso2022emergence}, nonnegativity and energy efficiency \citep{whittington2023disentanglement}, wiring costs \citep{blauch2022itn}, and smoothness \citep{margalit2024}. These studies do not, however, ask whether emergent selectivity can replace a deliberately designed receptive field. We address this comparison directly.
%%%%%%%%%%%%%%%%%%%%%%%%%%%%%%%%%%%%%%%%%%%%%%%%%
\section{Methods}
\label{sec:methods}
%%%%%%%%%%%%%%%%%%%%%%%%%%%%%%%%%%%%%%%%%%%%%%%%%
\subsection{Datasets and Sensory Coordinates}
We use three datasets spanning auditory and event-based visual classification, with a different task-relevant sensory coordinate in each case.

\paragraph{Audition.}
\emph{Spiking Heidelberg Digits (SHD)} \citep{shd} contains spoken digits (0--9, in English and German; 20 classes, chance 5\%) converted into spike trains by a cochlear model. Input channels correspond to frequency bands, making frequency the relevant one-dimensional sensory coordinate.
SHD requires minimal preprocessing because frequency creates explicit channel ordering (Appendix~\ref{sec:appendix-shd}). 

\paragraph{Event-based vision.}
\emph{DVS-Gesture} \citep{dvsgesture} contains recordings of 11 human gestures (chance $\approx$ 9.1\%) with a Dynamic Vision Sensor (DVS). Class identity is primarily encoded by motion, and we therefore identify motion direction as the relevant sensory coordinate.
\emph{CIFAR10-DVS} \citep{cifar10dvs} is generated by presenting CIFAR-10 images (10 classes, chance 10\%) to a DVS under artificial motion. Because this motion is shared across classes, object identity is primarily spatial, and the relevant coordinates are the two retinotopic axes.

In both DVS datasets, asynchronous events are accumulated into frames and binarized (Appendix~\ref{sec:appendix-dvs}). These frames directly provide the retinotopic coordinates used for CIFAR10-DVS; DVS-Gesture requires an additional step to introduce motion direction.

\paragraph{Motion preprocessing.}
For DVS-Gesture, we extract motion direction using a Hassenstein Reichardt-style correlator with null-direction opponency \citep{hassenstein1956,borst2015}. We apply it along the eight cardinal and diagonal directions, replacing retinotopic frames with direction channels (see Appendix~\ref{sec:appendix-motion}).

\subsection{The Expressive Leaky Memory Network}
All experiments use the Expressive Leaky Memory (ELM) network
\citep{spieler2026scaling}, which allows single neuron expressivity to be varied independently of connectivity. 
\paragraph{The ELM neuron.}
Each unit of the network is an ELM neuron \citep{spieler2024elm}, a
phenomenological neuron model designed to capture computation over multiple
timescales. Each ELM neuron carries $M$ memory units $m_t \in \mathbb{R}^M$:
internal variables that store information across time steps, each forgetting it
at its own fixed rate, or timescale $\tau_j$. Since every memory unit has a
distinct timescale, a neuron with $M$ memory units integrates information over
$M$ timescales that are spaced logarithmically over a fixed range. At each time step, a
small MLP combines the current synaptic input $s_t$ with the memory units to
compute an update,
\begin{equation}
m_t = \lambda \odot m_{t-1} + (1-\lambda) \odot \mathrm{MLP}\bigl([s_t, m_{t-1}]\bigr),
\qquad \lambda_j = e^{-\Delta t/\tau_j},
\label{eq:elm-neuron}
\end{equation}
and the neuron's output is read out from the updated memory $m_t$. In our
experiments, we vary $M$ at fixed network size $N$, and $N$ at fixed $M$.
\paragraph{Synaptic budget.}
Each neuron receives a fixed number of synapses $d_s$, of which a fraction $\rho_{rec}$ is allocated to recurrent connections and the remainder to feed-forward input. In the structured conditions this allocation is exact, with every neuron receiving $\mathrm{round}(d_s\rho_{rec})$ recurrent synapses. Our main manipulations change only how feed-forward synapses are distributed across the sensory input, holding $d_s$ and the expected value of $\rho_{rec}$ fixed across conditions.

The full input baseline is the only exception: each neuron connects to every input. Throughout, recurrent connections are drawn at random from $N$. Non-random recurrent topologies are examined separately in Appendix~\ref{sec:appendix-recurrent}.
\subsection{Experimental Conditions}
\label{sec:experimental_conds}
We compare two feed-forward input wiring conditions at matched parameter count:
\begin{figure}[htb]
    \floatconts
    {fig:methods0}
        {\caption{
            \textbf{Input wiring conditions.}
            Structured (top) and Random (bottom) input wiring on each dataset (columns). Left: hidden units receive feed-forward input synapses plus random recurrent synapses, and feed a readout, readout units project to class logits. Only the feed-forward wiring differs between rows. Colored boxes mark the inputs sampled by the hidden units (color matched); grey elements are unsampled.}
        }{\includegraphics[width=0.9\linewidth]{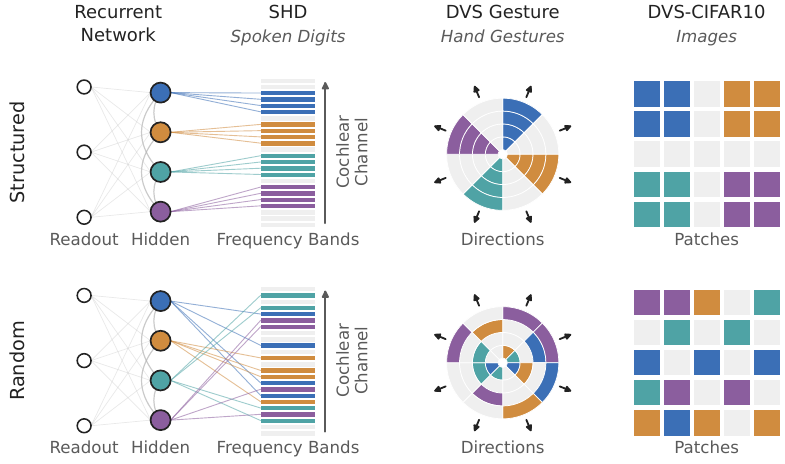}}
\end{figure}
\\
In the \emph{Structured Input Condition} each unit samples from a restricted region of the relevant sensory coordinate: neighboring frequency channels on SHD, one motion-direction channel on DVS-Gesture, or a spatial patch on CIFAR10-DVS. In the \emph{Random Input Condition} each unit samples uniformly from the full input array.
\\
Additional experiments use the \emph{Full Input Condition}, where each unit receives the entire input. We test this condition only on SHD, as the much larger event-camera inputs make it prohibitively expensive. All conditions are matched in training protocol.
\subsection{Statistical Analysis}
\label{sec:methods_statistics}
Each condition is run on 10 seeds, independently redrawing network initialization and connectivity. The unit of analysis is test accuracy from each trained network ($n=10$ per condition). Conditions are compared using two-sided Welch's $t$-tests, with multiple comparisons corrected using Benjamini--Hochberg. Appendix~\ref{sec:results_statistics} provides full statistics for all comparisons in the Results.

%%%%%%%%%%%%%%%%%%%%%%%%%%%%%%%%%%%%%%%%%%%%%%%%%
\section{Results}
\label{sec:results}
%%%%%%%%%%%%%%%%%%%%%%%%%%%%%%%%%%%%%%%%%%%%%%%%%
\subsection{Structured input improves performance but becomes less beneficial for expressive neurons.}
\label{sec:size_mem_sweep}
% --------------------
To test the effect of sensory-aligned input structure, we train ELM networks on SHD and DVS-Gesture with either structured or random feed-forward input wiring, aligned to frequency and motion direction, respectively. We repeat the comparison across network sizes and neuronal complexities, holding all other settings fixed within sweeps.
\begin{figure}[htbp]
    \floatconts
    {fig:results1}
{\caption{
\textbf{Structured input improves performance, with a reduced advantage at higher expressivity.}
Test accuracy for structured and random input wiring. Points show means across 10 seeds; shading shows s.e.m.
\\
\textbf{(A, B)} Structured input improves accuracy across all tested network sizes.
\\
\textbf{(C, D)} Increasing the number of memory units reduces this advantage. The conditions converge on SHD, while a smaller gap remains on DVS-Gesture.}
}{\includegraphics[width=\linewidth]{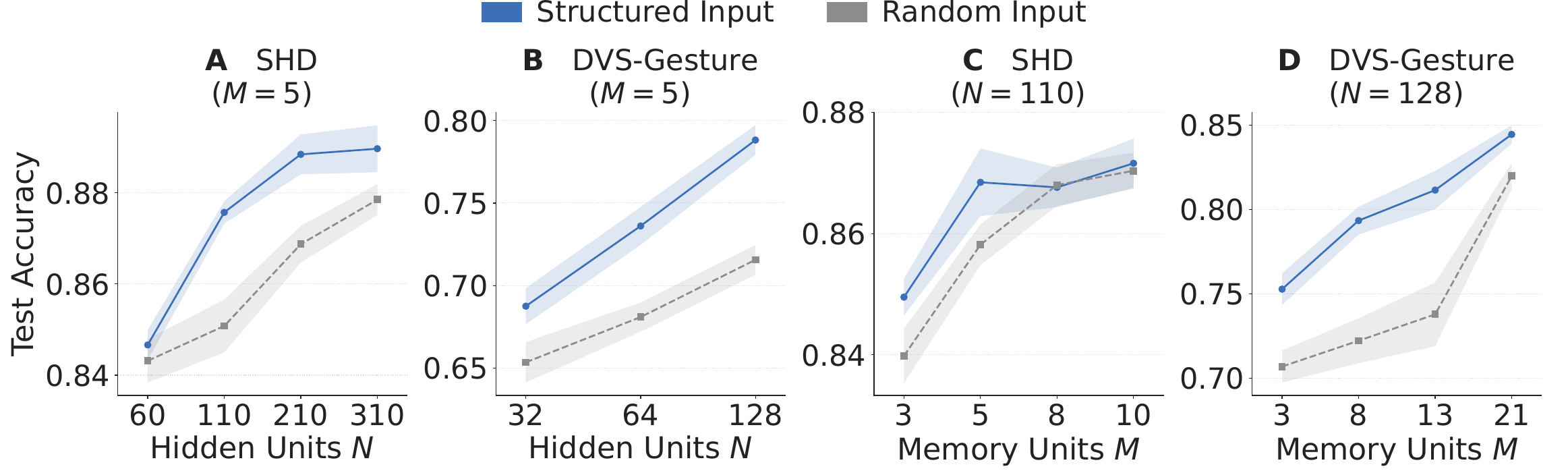}}
\end{figure}
Structured input improved test accuracy across all tested network sizes on both datasets (\figureref{fig:results1}\textbf{A}--\textbf{B}). Increasing neuronal complexity, however, reduced this advantage (\figureref{fig:results1}\textbf{C}--\textbf{D}). On SHD, structured and random input converged at the highest memory capacities tested. On DVS-Gesture, the gap also became substantially smaller at high memory capacity, although it did not fully disappear within the tested range. Thus, the benefit of structured input decreases as individual neurons become more expressive.

% --------------------
\subsection{Structured input helps only when aligned with task geometry.}
\label{sec:scrambled}
% --------------------
The two event-camera datasets allow us to test different receptive-field alignments on closely related input representations.
In DVS-Gesture, motion direction carries much of the information needed to distinguish gestures. 
In CIFAR10-DVS, however, motion is introduced only by the recording procedure and is shared across classes; object identity is instead encoded in spatial structure. We therefore compare motion-aligned and spatially structured input on both datasets. Motion-aligned input improves performance relative to random on DVS-Gesture but not on CIFAR10-DVS, while spatially structured input shows the opposite pattern (\figureref{fig:results2}\textbf{A}--\textbf{B}). So, structured input truly helps only when it is aligned with the sensory geometry relevant to the task.
\begin{figure}[htbp]
    \floatconts
    {fig:results2}
        {\caption{
            \textbf{Structured input helps only when aligned with task geometry.}
            Bars show means across 10 seeds; error bars show s.e.m. Brackets give BH-corrected Welch $t$-tests.
            \\
            \textbf{(A, B)} Motion-aligned input improves DVS-Gesture performance. Spatially structured input improves CIFAR10-DVS performance. Test accuracy is shown relative to the best condition.
            \\
            \textbf{(C, D)} Scrambling the sensory coordinate removes the structured-input advantage. Test accuracy is shown on an absolute scale.
            }
        }{\includegraphics[width=\linewidth]{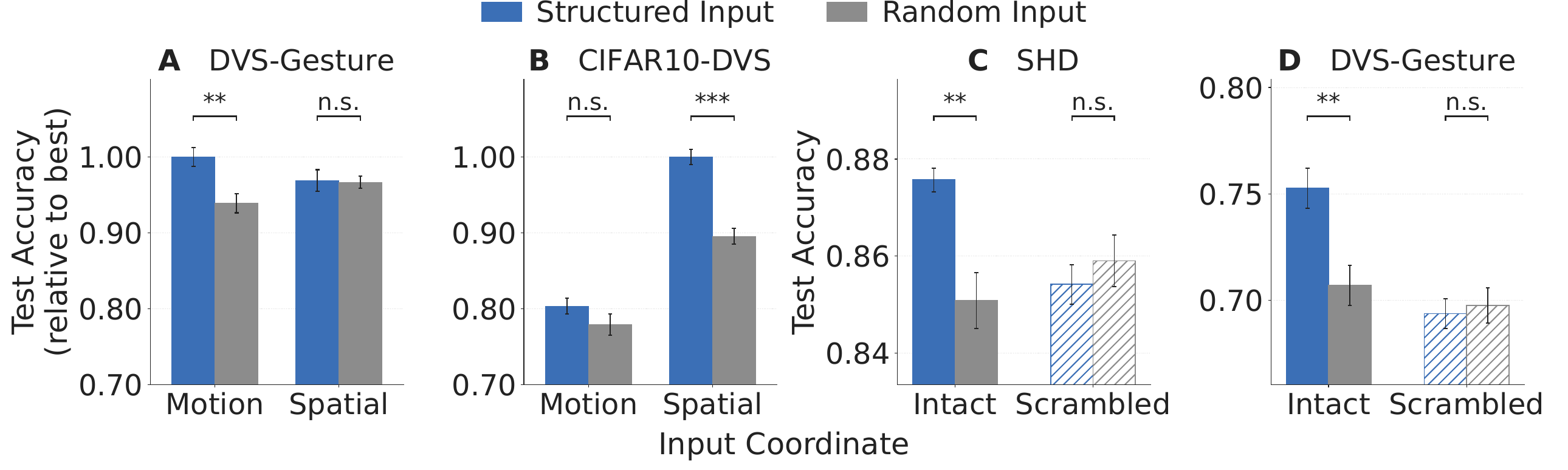}}
\end{figure}
As an additional control, we scramble the sensory coordinate before assigning structured input. This preserves the number and overlap of sampled inputs while removing their correspondence to neighboring positions in sensory space. The structured-input advantage then disappears on both SHD and DVS-Gesture (\figureref{fig:results2}\textbf{C}--\textbf{D}).

% --------------------
\subsection{Sparsity regularization partially recovers performance in full input networks.}
\label{sec:ff_reg_perf}
We next ask whether unrestricted access to the sensory input can compensate for the absence of receptive-field structure. On SHD, full input networks receive all input channels and therefore have substantially more feed-forward connections than the structured-input condition. 
\begin{figure}[htbp]
    \floatconts
    {fig:l1_improvement}
{\caption{
\textbf{Full input networks underperform structured input, and sparsity regularization partially recovers performance.}
All results are on SHD at $M=5$. Points show means across 10 seeds; shading and error bars show s.e.m.
\\
\textbf{(A, B)} Full input networks achieve lower test accuracy across network sizes and a larger train--test gap.
\\
\textbf{(C, D)} An $\ell_1$ penalty improves test accuracy and narrows the train--test gap, but does not reach structured-input performance.}
        }{\includegraphics[width=\linewidth]{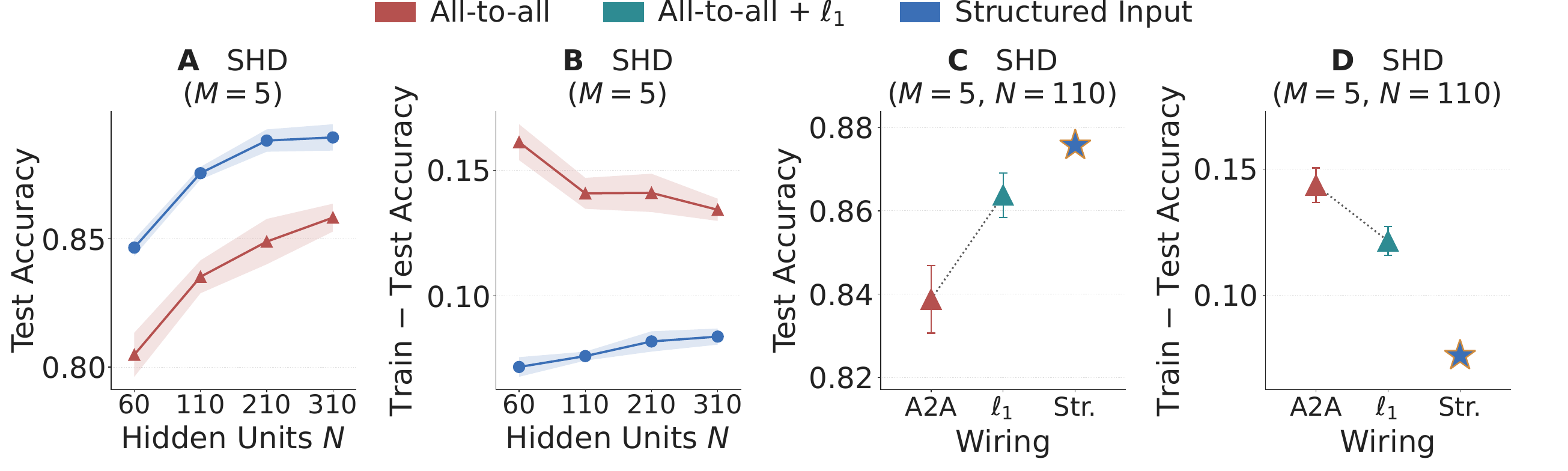}}
\end{figure}

Despite this additional capacity, they achieve lower test accuracy across network sizes and show a larger train--test gap, consistent with stronger overfitting (\figureref{fig:l1_improvement}\textbf{A}--\textbf{B}).

We therefore apply an $\ell_1$ penalty ($\lambda = 0.9$) to the feed-forward weights of the full input network (see Appendix~\ref{sec:appendix-regularization} for details). Regularization increases test accuracy and reduces the train--test gap, partially closing the performance difference to the structured-input condition, but does not eliminate it (\figureref{fig:l1_improvement}\textbf{C}--\textbf{D}).

In summary, full input provides more capacity but generalizes worse, while sparsity regularization only partially recovers the benefit of structured receptive fields.
% --------------------
\subsection{Sparsity regularization promotes input selectivity}
\label{sec:ff_reg_w}
Next, we examine how sparsity regularization changes feed-forward input weights. To quantify the emerging selectivity, we define the center of each putative receptive field as the input channel at which the cumulative absolute feed-forward weight reaches 50\% of the neuron's total input weight, and order neurons by this center. 

Without regularization, trained input weights remain broadly distributed across frequency channels after training (\figureref{fig:l1_emergence}\textbf{A}). With the penalty, most weights are driven to zero, and the remaining inputs concentrate within restricted frequency bands that vary across neurons (\figureref{fig:l1_emergence}\textbf{B}). 

The emerging receptive fields are concentrated in frequency ranges with substantial input activity and are largely absent where SHD contains few spikes (\figureref{fig:l1_emergence}\textbf{C}). They remain broad and overlapping, however, and do not recover the organization of the explicitly structured-input condition. 

\begin{figure}[htb]
\floatconts
{fig:l1_emergence}
    {
    \caption{
\textbf{$\ell_1$ regularization induces frequency-selective input structure.}
Feed-forward weights across SHD frequency channels for one representative seed, after training. Neurons are ordered by the center of their putative receptive field, defined as the channel at which cumulative absolute input weight reaches 50\% of the total (cyan).
\\
\textbf{(A)} No regularization, input weights are broadly distributed across frequency channels.
\\
\textbf{(B)} With regularization, most weights are driven to zero and the remaining inputs concentrate within frequency bands that vary across neurons; 26 neurons receive no feed-forward input.
\\
\textbf{(C)} Cumulative distribution of input spikes across SHD frequency channels.
}    }
{\includegraphics[width=0.9\linewidth]{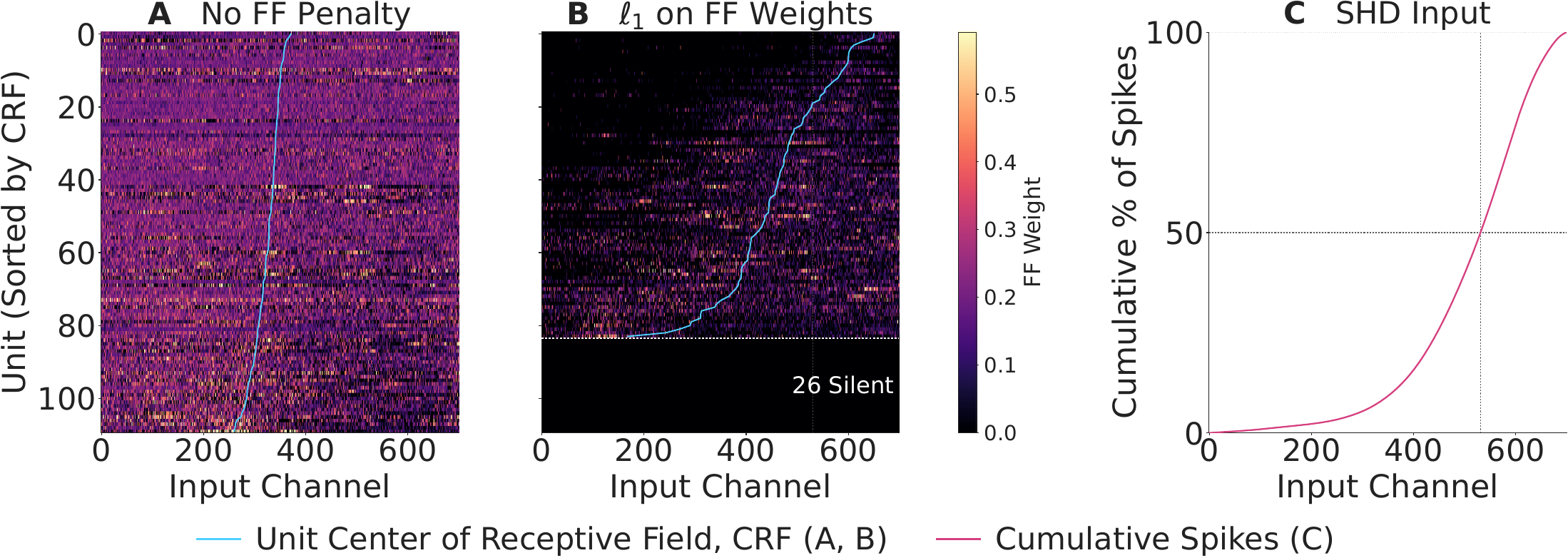}}
\end{figure}
%%%%%%%%%%%%%%%%%%%%%%%%%%%%%%%%%%%%%%%%%%%%%%%%%
\section{Discussion}
\label{sec:discussion}
%%%%%%%%%%%%%%%%%%%%%%%%%%%%%%%%%%%%%%%%%%%%%%%%%
Our results show that the proposed receptive-field structure can improve generalization when aligned with the sensory geometry of the task. Scrambling the sensory coordinates removes this advantage, indicating that connectivity alone is insufficient.
The benefit of this bias depends on neuronal expressivity. Increasing the number of memory units reduces the advantage of structured input, with convergence on SHD and a smaller gap on DVS-Gesture. Thus, simple units benefit more from task-aligned input selectivity, whereas expressive units can compensate for its absence.

Full input experiments distinguish this structural bias from generic regularization. Although full input networks have greater capacity, they generalize worse; $\ell_1$ regularization improves performance and induces coarse frequency-selective input fields, but does not fully recover the benefit of structured receptive fields. Sparsity therefore captures only part of the advantage.

From a biological perspective, our results address whether the receptive-field structure can be computationally useful, rather than how it emerges. Alignment with sensory geometry provides a computational advantage, although more expressive units can compensate for its absence. Our imposed feed-forward structure remains an abstraction of functional receptive fields and does not identify their biological mechanisms.

Finally, we considered only a small set of sensory geometries and tasks, with neuronal expressivity varied within a single model family. Testing other geometric priors and architectures will determine how general this trade-off is.
\newpage
\section*{Acknowledgments}
This work was funded by the Federal Ministry of Research, Technology and Space (BMFTR) through the RealmNN project under the NeuroAI initiative (grant number 01GQ2508).
We acknowledge support from the BMBF through the T\"ubingen AI Center (FKZ: 01IS18039B).
AL is a member of the Cluster of Excellence ``Machine Learning: New Perspectives for Science'', EXC 2064/1 (project number 39072764).

We thank the ML Cloud at T\"ubingen AI Center for providing computational resources on the Galvani high-performance computing system.

The authors declare no competing interests.
%%%%%%%%%%%%%%%%%%%%%%%%%%%%%%%%%%%%%%%%%%%%%%%%%
\bibliography{references}
%%%%%%%%%%%%%%%%%%%%%%%%%%%%%%%%%%%%%%%%%%%%%%%%%
\appendix
\section{Datasets, Preprocessing, and Network Configuration}
\label{sec:appendix}
\subsection{Spiking Heidelberg Digits}
\label{sec:appendix-shd}
The Spiking Heidelberg Digits \citep{shd} are spike-encoded recordings
of spoken digits (0--9, German or English), giving 20 classes (chance $= 5\%$). Each recording was converted from audio to spikes by a detailed model of the inner ear, yielding $C = 700$ binary input
channels ordered by the characteristic frequency of the cochlear section that drives them. Recordings are trimmed to one second and discretized into 2\,ms bins, producing $T = 500$ time steps.
\\
We use the official train and test files (8156 and 2264 recordings) and hold out a validation split by marking each training recording independently with probability 0.2, which yields roughly 6500 training and 1600 validation recordings. The draw uses a fixed seed that is independent of the training seed, so all seeds and all wiring conditions see exactly the same partition of the data. No augmentation is applied to SHD. The official test set holds out two speakers entirely, so absolute accuracies are lower than for a within-speaker split; this affects all conditions equally and does not enter any
of our comparisons.
\begin{figure}[htbp]
\floatconts
{fig:datasets}
{\caption{\textbf{The three datasets.}
  Representative samples from each dataset.
  In panels A and B each snapshot accumulates events over a short window centred on time $t_1$ or $t_2$.
  \\
  \textbf{(A)} Spiking Heidelberg Digits: one spoken English digit
  (``three''), rendered as a spike raster over the cochlear input
  channels. 
  \\
  \textbf{(B)} DVS1-28 Gesture: two snapshots from the class right-hand clockwise
  rotation.
  \\
  \textbf{(C)} CIFAR10-DVS: two snapshots of a sample from the
  \emph{cat} class.
}}
{\includegraphics[width=\linewidth]{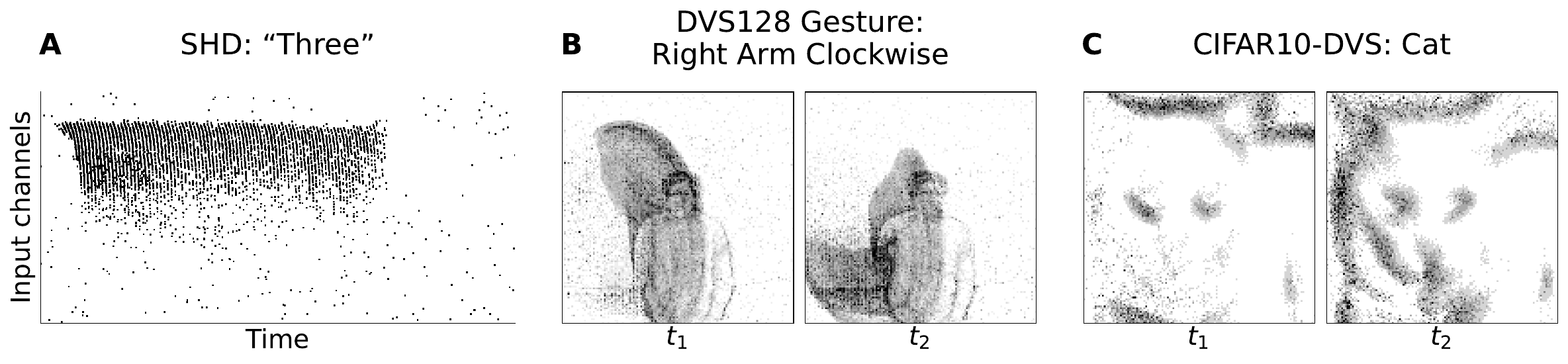}}
\end{figure}
\subsection{Event-Camera Datasets: DVS-Gesture and CIFAR10-DVS}
\label{sec:appendix-dvs}
Both event-camera datasets come as raw event streams: each recording
is a variable-length list of events $\{(t_k, x_k, y_k, p_k)\}_{k=1}^{N}$, where $(x_k, y_k)$ is the pixel that fired, $t_k$ the microsecond timestamp, and $p_k \in \{0,1\}$ the polarity, i.e.\ whether the log-intensity at that pixel rose (ON) or fell (OFF). The number of events $N$ differs from recording to recording, so the streams must be converted to fixed-size tensors before they
can be batched. 
\paragraph{Event Stream Processing} We do this with the frame-integration procedure of SpikingJelly \citep{spikingjelly}. Given a target frame count $T$, the event list of a recording, sorted by time, is partitioned into $T$ contiguous index intervals $[j_l^{(t)}, j_r^{(t)})$, under one of two splitting rules: \emph{by time},
where the recording duration $t_N - t_1$ is divided into $T$ equal spans and each interval collects the events falling in its span, or \emph{by number}, where each interval receives an equal count $\lfloor N/T \rfloor$ of consecutive events. Frame $t$ is then formed by counting, for each pixel and each polarity separately, how many events of interval $t$ landed there, giving a tensor of shape $T \times 2 \times H \times W$.  Pixel values at this stage are event counts, but they are binarized at the very end of preprocessing. The two rules differ in what they hold fixed: splitting by time preserves the true temporal scale, while
splitting by number equalizes how much evidence each frame carries and so adapts to recordings whose event rate varies.
\paragraph{DVS-Gesture} \cite{dvsgesture} contains recordings of 29 subjects performing 11 hand and arm gestures under three lighting conditions (chance $\approx 9.1\%$), captured with a $128 \times 128$ DVS event camera. We use the standard subject-wise split, 1176 training and 288 test recordings, and hold out 20\% of the training
recordings as validation with a fixed seed (42), again shared across all seeds and conditions. Each recording is integrated to $T = 100$ frames, split by time, so each frame spans an equal share of the gesture's duration. Frames are spatially binned by a factor of 2, giving a $64 \times 64$ grid from the native $128 \times 128$ sensor. The only augmentation is a random spatial translation of up to 8 pixels, drawn per sample and applied to the training split only; validation and test frames are never augmented.
\\
\paragraph{CIFAR10-DVS} \cite{cifar10dvs} was produced by displaying the 10000 static CIFAR-10 images on a monitor and recording each with a DVS camera while the image was moved along a repeating path, giving 10 classes with 1000 recordings each (chance $= 10\%$). The apparent motion in this dataset is therefore an artifact of the recording procedure rather than a property of the depicted object, which is
precisely why we include it: it is a vision dataset whose class-relevant structure is spatial, and it lets us ask whether the useful notion of alignment tracks the structure the task depends on. Recordings are integrated to $T = 20$ frames, split by number, since the repeating stimulus path makes event count the more stable unit here, and are binned by a factor of 2 to the same $64 \times 64$ grid with separate ON and OFF channels. We use a fixed random 8:1:1 train/validation/test partition (seed 42), as the dataset ships without an official split. No augmentation is applied.
\begin{table}[htbp]
\centering
\small
\caption{Dataset parameters.}
\label{tab:params_data}
\begin{tabular}{lccc}
\toprule
 & \textbf{SHD} & \textbf{DVS-Gesture} & \textbf{CIFAR10-DVS} \\
\midrule
Classes                          & 20   & 11   & 10 \\
Timesteps $T$                    & 500  & 100  & 20 \\
\addlinespace
\multicolumn{4}{l}{\emph{Input dimension $C$}} \\
\quad Spatial           & ---  & $2\times64\times64$  & $2\times64\times64$ \\
\quad Motion          & ---  & $16\times64\times64$ & $16\times64\times64$ \\
\quad Frequency         &700 & --- & --- \\
\bottomrule
\end{tabular}
\end{table}
\\
\indent Absolute accuracies on CIFAR10-DVS are relatively low (the best seeds reach $\sim40\%$ test accuracy). We train for 50 epochs on 10,000 recordings and do not optimize the experimental setup for absolute performance, as the dataset is used here only as a case where the class-relevant coordinate is spatial rather than motion-based. Our focus is therefore on relative performance, and we report accuracies normalized by the best accuracy achieved in our experiments.
\\
We compare each structured condition only against its own budget-matched random control, and never motion against spatial directly. The
two coordinates present the network with different input spaces: the motion
condition sees all 16 direction-polarity channels, whereas the spatial
condition sees the $64 \times 64 \times 2$ retinotopic grid for 2 polarities. A
direct contrast between them would confound the sensory coordinate with the
size and content of the input array, so we do not report one.
\\
Both event datasets are preprocessed once to frame tensors and cached to disk. 
\subsection{Motion Preprocessing: The Direction-Selective Correlator}
\label{sec:appendix-motion}

The frame sequence is decomposed into direction-selective channels by a
Hassenstein--Reichardt-style correlator with null-direction opponency
\citep{hassenstein1956,borst2015}. Motion direction is read from the order in
which neighboring pixels become active: a rightward stimulus reaches the left
neighbor first and the pixel itself a moment later.
\begin{figure}[htbp]
\floatconts
{fig:correlator}
{\caption{Direction selectivity through opponency.}}
{\includegraphics[width=0.5\linewidth]{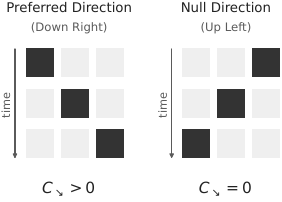}}
\end{figure}

Each channel compares two opposite neighbors of a pixel: the \textbf{preferred}
neighbor $p - \delta(\theta)$, from which motion in direction $\theta$ would
have arrived, and the \textbf{null} neighbor $p + \delta(\theta)$ on the
opposite side. It correlates the pixel's current activity $s_p(t)$ with an
exponential moving average of each neighbor's recent activity,
$M_p(t) = \lambda M_p(t-1) + (1-\lambda)\,s_p(t-1)$, and rectifies the opponent
difference:
\begin{equation}
C_\theta(p,t) = \mathrm{ReLU}\!\left[\,
s_p(t)\bigl(M_{p-\delta(\theta)}(t) - M_{p+\delta(\theta)}(t)\bigr)\right].
\label{eq:hr-correlator}
\end{equation}
Preferred-direction motion makes the difference positive; null-direction motion
activates the same pixels in reverse order, making it negative, and the
rectifier sets $C_\theta = 0$ (Figure~\ref{fig:correlator}). The difference also
cancels when both sides are equally active, and the $s_p(t)$ factor keeps the
channel silent wherever the pixel itself is not active.

We set $\lambda = 0.6$, an effective horizon of $1/(1-\lambda) = 2.5$ frames,
with $M_p(0) = 0$ and missing neighbors at the grid boundary set to zero. Each
correlator compares a pixel with a neighbor one pixel away in the cardinal
directions and $\sqrt{2}$ pixels away in the diagonal directions. The correlator
runs at every pixel for the four cardinal and four diagonal directions
($|\Theta| = 8$, spaced $45^\circ$), separately on the ON and OFF streams,
expanding the channel count from 2 (original ON and OFF) to 16.

\subsection{The Expressive Leaky Memory Network}
\label{sec:appendix-parameters}
\paragraph{The ELM network.}
We use the ELM Network of \cite{spieler2026scaling}: a doubly recurrent model
whose core is a wide recurrent layer in which each unit is itself recurrent. The
hidden layer holds $N_{\mathrm{hidden}}$ independently parameterized ELM units.
At each timestep the layer forms the concatenation of its feed-forward input
$u_t$ with its own previous activity $a_{t-1}$, and every unit reads a fixed
subset of $d_s$ channels from it, assigned once at initialization:
\begin{equation}
z_{t,i} = \mathrm{input\_select}_i\bigl([\,u_t,\; a_{t-1}\,]\bigr),
\qquad
a_{t,i} = \mathrm{ELM}_i(z_{t,i}).
\label{eq:elm-layer}
\end{equation}
Because each unit has its own parameters and its own input subset, both the
per-unit synapse budget $d_s$ and the fraction of those synapses that are
recurrent are set explicitly. In \cite{spieler2026scaling} that subset is drawn
uniformly at random, with $\rho_{\mathrm{rec}}$ the probability of sampling a
recurrent rather than a feed-forward connection. The sampling of
$\mathrm{input\_select}_i$ is our experimental variable: we vary how each unit's
channels are chosen while holding $d_s$ and $\rho_{\mathrm{rec}}$ fixed, so the
conditions are matched in parameter count and differ only in which channels each
unit sees. The wiring schemes are described below (Section~\ref{sec:appendix-wiring}).

The hidden layer is followed by a smaller feed-forward readout layer of simpler
ELM units and a final linear projection to the class logits.
All parameters are listed in Table~\ref{tab:params_network}.
\begin{table}[htbp]
\centering
\small
\caption{Network parameters, given for the SHD reference configuration. The
network is specified once and reused across datasets and wiring conditions;
only the entries marked $\dagger$ vary in our experiments.}
\label{tab:params_network}
\begin{tabular}{ll}
\toprule
\multicolumn{2}{l}{\textbf{Hidden layer}} \\
Hidden units $N_{\mathrm{hidden}}$ & 110$^\dagger$ \\
Memory traces $M$ & 5$^\dagger$ \\
Memory $\tau$ range & $[0.1, 100]$ \\
Timestep $\Delta t$ & 1.0 \\
Branches $B$ & 21$^\dagger$ \\
Synapses per branch $d_s/B$ & 4$^\dagger$ \\
Per-unit budget $d_s$ & 84$^\dagger$ \\
Recurrent fraction $\rho_{\mathrm{rec}}$ & 0.125 \\
MLP hidden layers & 1 \\
MLP hidden units & $2M = 10$ \\
MLP activation & ReLU$^2$ \\
Unit activation & ReLU \\
Synapse rectifier & ReLU \\
Synapse weight init & 1.0 (constant) \\
Synapse scale & 1.0 \\
Input activation gain $g$ & 10.0 \\
Input embedding & scaled, factor 1.0 \\
Input dropout & 0.2 \\
Recurrent dropout & 0.2 \\
\midrule
\multicolumn{2}{l}{\textbf{Readout layer}} \\
Wiring & all-to-all  \\
Memory traces $M$ & 3 \\
Branches $\times$ synapses per branch & $22\times5 = 110$$^\dagger$ \\
MLP hidden layers & 0 \\
Unit activation & identity \\
Output embedding & linear layer \\
\bottomrule
\end{tabular}
\end{table}

\subsection{Wiring Conditions}
\label{sec:appendix-wiring}

A wiring condition is a rule assigning each hidden unit its list of $d_s$
synaptic sources, drawn from the concatenation of the $C$ sensory channels
(feed-forward) and the $N_{\mathrm{hidden}}$ hidden units at the previous
timestep (recurrent). Lists are drawn once at initialization and fixed thereafter.
The conditions differ only in the feed-forward assignment: recurrent synapses are
always drawn uniformly from the full hidden population. In the structured
conditions the per-unit counts are fixed,
$d_{\mathrm{rec}} = \mathrm{round}(d_s \rho_{\mathrm{rec}})$ and
$d_{\mathrm{ff}} = d_s - d_{\mathrm{rec}}$, whereas the random condition fixes
them only in expectation.

\paragraph{Random input.} The control condition. Every synapse is drawn
independently: with probability $\rho_{\mathrm{rec}}$ its source is a hidden
unit, drawn uniformly from all $N_{\mathrm{hidden}}$, and otherwise a sensory
channel, drawn uniformly from all inputs. The feed-forward/recurrent split therefore
holds in expectation, each unit drawing
$\mathrm{Binomial}(d_s, \rho_{\mathrm{rec}})$ recurrent sources rather than
exactly $\rho_{\mathrm{rec}} d_s$, so synaptic composition varies across units.

\paragraph{Structured input, frequency coordinate (SHD).} The $C=700$ input channels are divided into 10 contiguous windows along the frequency axis, with a $5\%$ overlap between neighboring windows. Units are assigned to windows in index order and connect to all channels within their window, resulting in windows of 74 channels and $d_{\mathrm{ff}}=74$. We set the synapse budget to $d_s=84$ accordingly.

\paragraph{Structured input, motion coordinate (DVS-Gesture, CIFAR10-DVS).} The
input comprises $\text{num\_channels} = 16$ direction-polarity channels, each on a $64 \times 64$ grid. The hidden population is partitioned into $16$ equal
clusters, one per channel, and a unit in cluster $c$ draws its $d_{\mathrm{ff}}$
synapses uniformly from the $64 \times 64$ positions of channel $c$. The
constraint thus acts on the direction coordinate alone, leaving retinotopic
position unconstrained.

\paragraph{Structured input, spatial coordinate (DVS-Gesture, CIFAR10-DVS).} The input is the $64 \times 64$ retinotopic grid of each polarity, partitioned
into a $4 \times 4$ array of non-overlapping patches. Each unit is assigned one
patch and draws all $d_{\mathrm{ff}}$ synapses uniformly within it. The
constraint here acts on the two retinotopic coordinates and leaves direction
unconstrained, the complement of the motion condition.

\paragraph{Scrambled control.} Scrambling alters the \emph{input}, not the
wiring. A single fixed permutation $\pi$ of the sensory coordinate is drawn once and applied identically to the training, validation
and test splits: on SHD it permutes the $700$ channel indices, on the event
datasets the $64 \times 64$ positions within each channel before the motion decomposition is performed. The wiring rules above are then applied unchanged, to both the
structured and the random input conditions. Each unit therefore still draws the same
number of sources with the same overlap; what is removed is the correspondence
between those indices and neighboring positions on the sensory coordinate.
\subsection{Regularization}
\label{sec:appendix-regularization}
Two regularizers are active in every network we report, with identical
strengths across wiring conditions and datasets:
\begin{itemize}
\item \textbf{Neuron MLP weights} ($\ell_2$, strength $0.01$). Applied per unit
to the mean absolute weight of its internal MLP.
\item \textbf{Activity} ($\ell_1$, strength $1.0$). Applied to the mean
population activity of the layer.
\end{itemize}
A third regularizer, a \textbf{feed-forward weight} penalty ($\ell_1$, strength
$0.9$, applied per unit to the ReLU-rectified feed-forward weights), is applied only in the subset of
full input experiments reported in Sections~\ref{sec:ff_reg_perf}
and~\ref{sec:ff_reg_w}. It is introduced gradually with a sigmoid schedule that reaches its maximum halfway through training. 
\paragraph{Sensitivity to penalty strength.} 
\begin{figure}[htbp]
\floatconts
{fig:reg_sweep}
{\caption{\textbf{Feed-forward weights across $\ell_1$ penalty strengths.} Single run per
strength (SHD, $N = 110$, $M = 5$). Units are sorted by receptive-field center (cyan;
defined as in Figure~5); the dotted vertical line marks the median channel of SHD input
spikes, and units below the horizontal line receive no feed-forward input.}}
{\includegraphics[width=\linewidth]{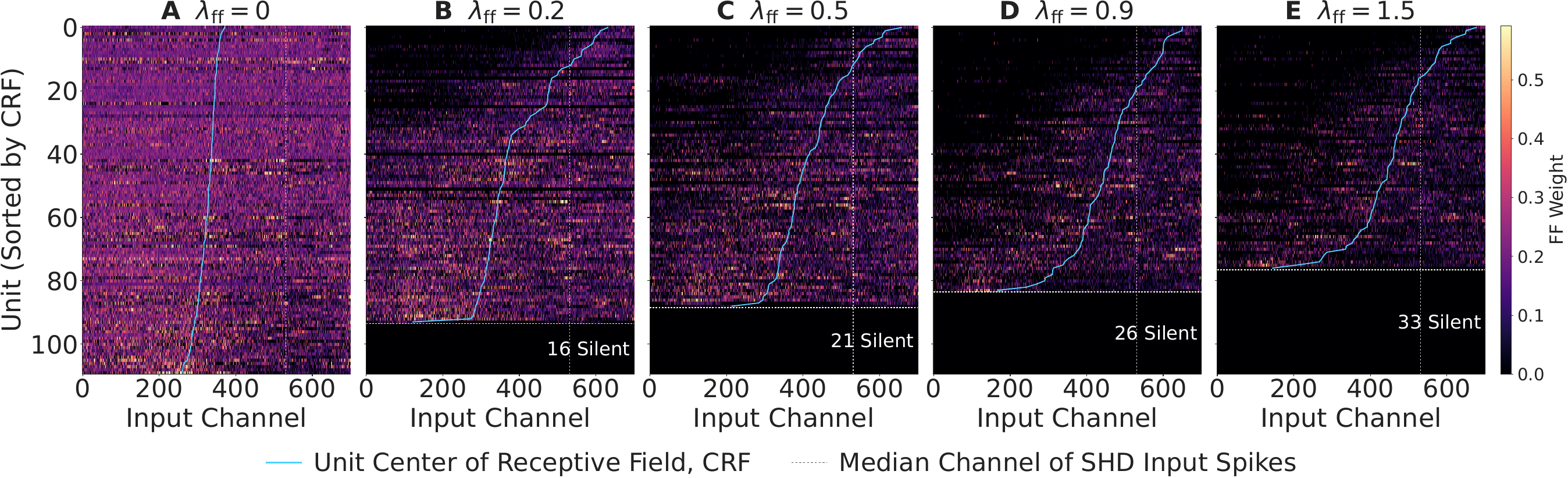}}
\end{figure}
The results in Sections~\ref{sec:ff_reg_perf} and~\ref{sec:ff_reg_w} use a fixed feed-forward penalty strength $\lambda_{\mathrm{ff}} = 0.9$. To assess the sensitivity of the results to this choice, we train one additional seed for each $\lambda_{\mathrm{ff}} \in \{0.2, 0.5, 1.5\}$. Test accuracy was higher than for the unregularized baseline at all three strengths. Given the seed-to-seed variability observed in previous experiments, these single runs do not support definite conclusions about relative performance across penalty strengths. However, the learned feed-forward weight patterns were consistent across seeds in the previous experiments, allowing us to draw qualitative conclusions from these additional runs (Figure ~\ref{fig:reg_sweep}). Increasing regularization promotes sparsity, with approximately 80\% of feed-forward weights zero at the highest strength tested. Nevertheless, none of the regularization strengths tested recovers the receptive field structure imposed by the structured-input condition; instead, the remaining input fields stay broad and overlapping.
\subsection{Training and Optimization Settings}
\label{sec:appendix-optimization}
For all three datasets, the network processes the full input sequence and predicts the class from the logits at the final timestep, using a standard cross-entropy loss.
\\
Table~\ref{tab:optim} lists the optimizer settings.
\begin{table}[htbp]
\centering
\small
\caption{Optimization settings. Identical across wiring conditions and, except
for the epoch count, across datasets.}
\label{tab:optim}
\begin{tabular}{ll}
\toprule
Optimizer & Adamax \\
Learning rate & $5\times10^{-3}$, cosine decay to 0 \\
Warmup & 400 steps, linear \\
Batch size (train / eval) & 8 / 8 \\
Gradient clipping & global norm 1.0 \\
Label smoothing & 0.01 \\
Epochs & 100 (SHD), 150 (DVS-Gesture), 50 (CIFAR10-DVS) \\
Seeds per condition & 10 \\
\bottomrule
\end{tabular}
\end{table}
\\
The network-size and memory sweeps comprise independent sets of runs and differ in one protocol detail. The size sweep uses early stopping on validation accuracy, stopping after 20 epochs without an improvement of at least 0.01, whereas the memory sweep is trained for the full epoch budget. This prevents training duration from becoming a confound in the memory sweep, allowing it to isolate the effect of memory capacity. Data-split seeds are held fixed and independent of the training seed.

\section{Statistical Analysis of the results}
\label{sec:results_statistics}
% --------------------

Table~\ref{tab:appendix-stats} lists every comparison behind the Results. Each comparison contrasts two independent sets of $n = 10$ trained networks, one per wiring condition, where the unit of analysis is the test accuracy of a single network and each seed redraws both the network initialization and the connectivity. Because the two conditions need not have equal variance, every comparison is a two-sided Welch's $t$-test on those two sets of $10$ accuracies. 
\\
Each row of the table gives the difference between the two conditions in
test-accuracy points, signed so that a positive value favors the first-named arm,
and the Benjamini--Hochberg $q$-value. A comparison is called significant
only if it meets $q < 0.05$.
\\
We apply multiple-comparison correction across all 39 reported tests, including the repeated contrasts that appear in multiple blocks. Three contrasts are repeated because they address different questions. However, correcting across the 36 unique comparisons yields the same 25 significant results, with $q$-values differing only in the fourth decimal place.
\\
For completeness, we also computed the correction within each of six
subsets, one per scientific question, defined before the pooled analysis was
run. That partition yields the identical set of significant comparisons, so
no reported conclusion turns on the choice.
\begin{table}[htbp]
\setlength{\tabcolsep}{3pt}
\centering
\footnotesize
\caption{Statistical comparisons behind the Results}
\label{tab:appendix-stats}
\begin{tabular}{llrr}
\toprule
Comparison & Condition & $\Delta$ (pts) & $q$ \\
\midrule

\multicolumn{4}{l}{\textit{Network size sweep: structured $-$ random} (\figureref{fig:results1}\textbf{A}--\textbf{B})} \\
SHD (M=5)         & $N=60$   & $+0.35$ & $0.618$ \\
             & $N=110$  & $+2.49$ & $\mathbf{0.0042}$ \\
             & $N=210$  & $+1.97$ & $\mathbf{0.0071}$ \\
             & $N=310$  & $+1.12$ & $0.122$ \\
DVS-Gesture (M=5)  & $N=32$   & $+3.40$ & $0.079$ \\
             & $N=64$   & $+5.52$ & $\mathbf{0.0039}$ \\
             & $N=128$  & $+7.26$ & $\mathbf{<0.001}$ \\
\midrule
\multicolumn{4}{l}{\textit{Memory sweep: structured $-$ random} (\figureref{fig:results1}\textbf{C}--\textbf{D})} \\
SHD (N =110)          & $M=3$    & $+0.97$ & $0.122$ \\
             & $M=5$    & $+1.03$ & $0.175$ \\
             & $M=8$    & $-0.04$ & $0.934$ \\
             & $M=10$   & $+0.12$ & $0.852$ \\
DVS-Gesture  (N=128) & $M=3$    & $+4.58$ & $\mathbf{0.0058}$ \\
             & $M=8$    & $+7.12$ & $\mathbf{0.0014}$ \\
             & $M=13$   & $+7.36$ & $\mathbf{0.0079}$ \\
             & $M=21$   & $+2.47$ & $\mathbf{0.031}$ \\
\midrule
\multicolumn{4}{l}{\textit{Alignment: structured $-$ its random control} (\figureref{fig:results2}\textbf{A}--\textbf{B})} \\
DVS-Gesture ($N=128$, $M=3$) & motion   & $+4.58$ & $\mathbf{0.0058}$ \\
             & spatial  & $+0.17$ & $0.913$ \\
CIFAR10-DVS ($N=128$, $M=3$)  & motion   & $+0.93$ & $0.235$ \\
             & spatial  & $+4.02$ & $\mathbf{<0.001}$ \\
\midrule
\multicolumn{4}{l}{\textit{Scrambling control} (\figureref{fig:results2}\textbf{C}--\textbf{D})} \\
SHD ($N=110$, $M=5$) & struct $-$ random, intact    & $+2.49$ & $\mathbf{0.0042}$ \\
             & struct $-$ random, scrambled & $-0.49$ & $0.545$ \\
             & structured, intact $-$ scrambled & $+2.16$ & $\mathbf{0.0014}$ \\
             & random, intact $-$ scrambled & $-0.82$ & $0.375$ \\
DVS-Gesture ($N=128$, $M=3$) & struct $-$ random, intact    & $+4.58$ & $\mathbf{0.0058}$ \\
             & struct $-$ random, scrambled & $-0.38$ & $0.792$ \\
             & structured, intact $-$ scrambled & $+5.90$ & $\mathbf{<0.001}$ \\
             & random, intact $-$ scrambled & $+0.94$ & $0.545$ \\
\midrule
\multicolumn{4}{l}{\textit{Full input $-$ structured input, SHD M=5} (\figureref{fig:l1_improvement}\textbf{A}--\textbf{B})} \\
Test accuracy & $N=60$  & $-4.18$ & $\mathbf{0.0022}$ \\
              & $N=110$ & $-4.05$ & $\mathbf{<0.001}$ \\
              & $N=210$ & $-3.94$ & $\mathbf{0.0039}$ \\
              & $N=310$ & $-3.13$ & $\mathbf{0.0018}$ \\
Train--test gap & $N=60$  & $+8.92$ & $\mathbf{<0.001}$ \\
              & $N=110$ & $+6.46$ & $\mathbf{<0.001}$ \\
              & $N=210$ & $+5.90$ & $\mathbf{<0.001}$ \\
              & $N=310$ & $+5.03$ & $\mathbf{<0.001}$ \\
\midrule
\multicolumn{4}{l}{\textit{$\ell_1$ regularization, SHD $N=110$, $M=5$} (\figureref{fig:l1_improvement}\textbf{C}--\textbf{D})} \\
Regularized $-$ unregularized & test  & $+2.50$ & $\mathbf{0.034}$ \\
                              & gap   & $-2.21$ & $\mathbf{0.035}$ \\
Regularized $-$ structured    & test  & $-1.19$ & $0.092$ \\
                              & gap   & $+4.54$ & $\mathbf{<0.001}$ \\
\bottomrule
\end{tabular}
\end{table}
\section{Recurrent Wiring Experiments}
\label{sec:appendix-recurrent}
% --------------------

All structured conditions in the main text restrict only the feed-forward
connections. Recurrent sources are always drawn uniformly from the hidden
population, so the recurrent connectivity is statistically identical to the
random-input control. The receptive-field prior therefore acts on what a unit
reads from the sensory array, and not on which other units it reads from. Here
we ask whether extending the same structured organization to the recurrent draw adds
anything.
\\
We test three alternative recurrent samplers. All three leave the feed-forward
connections and the total synapse count $d_s$ unchanged, and replace only the
distribution from which each unit draws its $d_{\mathrm{rec}}$ recurrent
sources.
% --------------------
\paragraph{Strict modular recurrence.} Units are partitioned into modules by
their selectivity, using the same partition that already defines their
feed-forward windows: one module per frequency window on SHD, one module per
direction channel on the event datasets. A unit draws its recurrent sources only
from inside its own module. The recurrent connectivity is therefore block
diagonal and the modules are mutually disconnected.
% --------------------

% --------------------
\paragraph{Flexible modular recurrence.} The same partition with the hard
constraint relaxed. A fraction $1 - p_{\mathrm{inter}}$ of each unit's recurrent
sources lies inside its own module, and the remaining
$p_{\mathrm{inter}} = 0.1$ is spread uniformly over all units outside it.
Setting $p_{\mathrm{inter}} = 0$ recovers the strict condition, and
$p_{\mathrm{inter}} = 1 - m/N_{\mathrm{hidden}}$, with $m$ the module size,
recovers the uniform draw of the structured condition.
% --------------------

% --------------------
\paragraph{Small-world recurrence.} Units are placed on a ring ordered by
selectivity, and each unit is given a neighborhood  of the $k$ ring positions on
either side of it. A fraction $1 - p_{\mathrm{rewire}}$ of its recurrent sources
is drawn uniformly from inside that neighborhood ,and the remaining
$p_{\mathrm{rewire}} = 0.1$ from the positions outside it, which act as
shortcuts. The ring wraps, so the ordering is cyclic. We set $k$ so that the
neighborhood  is one module wide, $k = N_{\mathrm{hidden}} /
\text{num\_channels}$.
\\
The ring order is where the sensory coordinate enters, and it differs by
dataset. On SHD the ordering is the frequency axis. Unit index already
determines the feed-forward window, so consecutive ring positions are tuned to
neighbouring frequency bands. On the event datasets the ordering is angular. We
sort the direction channels by preferred angle and lay them on the ring in that
order, so ring neighbors are tuned to directions $45^{\circ}$ apart.
% --------------------

% --------------------
\paragraph{Recurrent structure does not change performance.} Across datasets and
network sizes we observed no consistent improvement relative to uniform
recurrence. 

Modular and small-world connectivity is typically reported to \emph{emerge} when units are
embedded in a metric space and the objective penalizes wiring length
\citep{chklovskii2002,blauch2022itn,margalit2024}. Under such a constraint,
clustering and shortcut topologies are efficient, because they achieve short
average path length at low wiring cost. Our units have no spatial embedding. Imposing a connectivity pattern that is optimal under a wiring
budget, in a model that has no wiring budget, adds a constraint without the
pressure that would make it beneficial.
\\
This also delimits the claim in the main text. The benefit we report comes from
aligning a unit's feed-forward inputs with the sensory geometry of the task, and
not from the arrangement of units relative to one another. It is consistent with
the interpretation in Section~\ref{sec:discussion}: the structured prior is
useful as it supplies a unit with the inputs it needs to compute over,
which is precisely what sufficiently expressive units can compensate for.
% --------------------

% --------------------
\section{Data and Code Availability}
\label{sec:data_code_aval}
SHD \citep{shd} is available from Zenke Lab at
\url{https://zenkelab.org/resources/spiking-heidelberg-datasets-shd/} under the
Creative Commons Attribution 4.0 International License (CC BY 4.0). DVS-Gesture \citep{dvsgesture} is available from IBM Research at \url{https://research.ibm.com/interactive/dvsgesture/} under CC BY 4.0. CIFAR10-DVS \citep{cifar10dvs} is available from figshare at
\url{https://figshare.com/articles/dataset/CIFAR10-DVS_New/4724671} under CC BY 4.0.
\\
Raw AEDAT event streams are integrated into frame tensors using SpikingJelly
\citep{spikingjelly}, available at
\url{https://github.com/fangwei123456/spikingjelly} under the Apache License
2.0, as described in Section~\ref{sec:appendix-dvs}; the event-camera
dataloaders read off the frame tensors directly.
\\
All code needed to reproduce the experiments, including the wiring conditions,
the event-camera dataloaders and the experiment configurations, is available at
\url{https://github.com/AgneseAdorante/elmnetwork}.

%%%%%%%%%%%%%%%%%%%%%%%%%%%%%%%%%%%%%%%%%%%%%%%%%
\end{document}